\documentclass[10pt,twocolumn,letterpaper]{article}

\usepackage{cvpr}
\usepackage{times}
\usepackage{epsfig}
\usepackage{graphicx}
\usepackage{amsmath}
\usepackage{amssymb}

\usepackage{multirow}

\usepackage{algorithm}
\usepackage{algorithmic}

\usepackage[pagebackref=true,breaklinks=true,letterpaper=true,colorlinks,bookmarks=false]{hyperref}

\cvprfinalcopy 

\def\cvprPaperID{0490} 

\ifcvprfinal\fi
\begin{document}

\title{Mining Deep And-Or Object Structures via Cost-Sensitive Question-Answer-Based Active Annotations}

\author{Quanshi Zhang$^{\dag}$, Ying Nian Wu$^{\ddag}$, Hao Zhang$^{\dag,\|}$, Song-Chun Zhu$^{\ddag}$\\
$^{\dag}$Shanghai Jiao Tong University,\quad$^{\ddag}$University of California, Los Angeles,\\$^{\|}$Chongqing University of Posts and Telecommunications}

\maketitle

\begin{abstract}
This paper presents a cost-sensitive active Question-Answering (QA) framework for learning a nine-layer And-Or graph (AOG) from web images\footnote[1]{Quanshi Zhang is the corresponding author. Quanshi Zhang is with the John Hopcroft Center and the MoE Key Lab of Artificial Intelligence, AI Institute, Shanghai Jiao Tong University. Ying Nian Wu and Song-Chun Zhu are with the Center for Vision, Cognition, Learning, and Autonomy, University of California, Los Angeles. Hao Zhang is with Chongqing University of Posts and Telecommunications. This work was done when Hao Zhang was an internship student at the Shanghai Jiao Tong University.}. The AOG explicitly represents object categories, poses/viewpoints, parts, and detailed structures within the parts in a compositional hierarchy. The QA framework is designed to minimize an overall risk, which trades off the loss and query costs. The loss is defined for nodes in all layers of the AOG, including the generative loss (measuring the likelihood of the images) and the discriminative loss (measuring the fitness to human answers). The cost comprises both the human labor of answering questions and the computational cost of model learning. The cost-sensitive QA framework iteratively selects different storylines of questions to update different nodes in the AOG. Experiments showed that our method required much less human supervision (\emph{e.g.} labeling parts on 3--10 training objects for each category) and achieved better performance than baseline methods.
\end{abstract}

\section{Introduction}

\subsection{Motivation \& objective}

Image understanding is one of core problems in the field of computer vision. Compared to object-detection techniques focusing on the ``what is where'' problem, we are more interested in mining the semantic hierarchy of object compositions and exploring how these compositions/sub-compositions are organized in an object. Such knowledge is a prerequisite for high-level human-computer dialogue and interactions in the future.


Therefore, in this paper, we aim to mine deep structures of objects from web images. More importantly, we present a cost-sensitive active Question-Answering (QA) framework to learn the deep structure from a very \textit{limited number} of part annotations. Our method has the following three characteristics.

\textbf{Deep and transparent representation of object compositions:} In fact, obtaining a \textit{transparent representation of the semantic hierarchy} is equivalent to \textit{understanding detailed object statuses}, to some extent. Based on such a hierarchical representation, parsing an entire object into different semantic parts and aligning different sub-components within each part can provide rich information in object statuses, such as the global pose, viewpoint, and local deformation of each certain part.

\begin{figure}
\centering
\includegraphics[width=\linewidth]{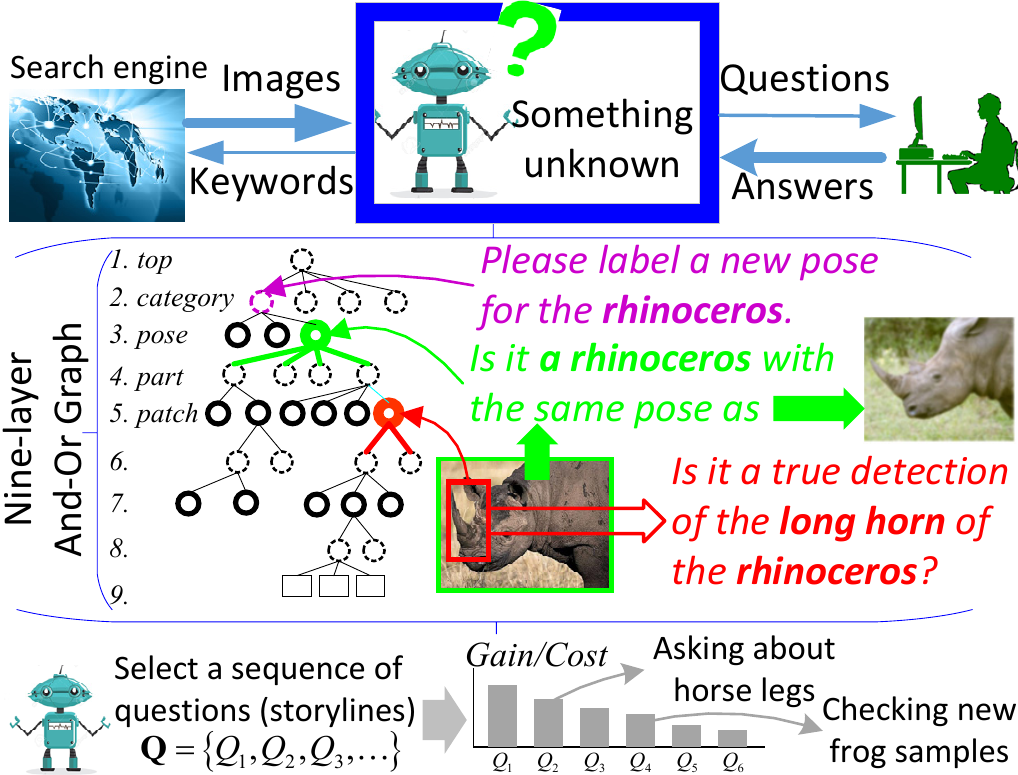}
\caption{Active QA framework. The QA framework automatically collects web images from the internet, selects something unknown to ask human beings, and uses the answer to learn an And-Or graph (AOG). The AOG represents deformable structures and compositional relationships between semantic visual concepts using a 9-layer hierarchy (see Fig.~\ref{fig:AOG}). We formulate the generative loss and discriminative loss for the AOG, and design different questions (see Fig.~\ref{fig:QAs}). Each question is used to refine a certain node in the AOG. The QA framework trades off the cost and potential gain (loss decrease) of each question, and selects the best sequence of questions.}
\label{fig:top}
\end{figure}

Thus, as shown in Fig.~\ref{fig:top}, a nine-layer And-Or graph (AOG) is proposed to represent visual concepts at different layers that range from \textit{categories}, \textit{poses/viewpoints}, \textit{parts}, to \textit{shape primitives} with clear semantic meanings. In the AOG, an AND node represents sub-region compositions of a visual concept, and an OR node lists some alternative appearance patterns of the same concept. Unlike modeling visual contexts and taxonomic relationships at the \textit{object} level in previous studies, the AOG focuses on semantic object components and their spatial relationships.

\textbf{Multiple-shot QA learning from big data:} In order to scale up the technique to big data, we apply the following two strategies to limit the annotation cost. First, we collect web images using search engines as training samples without annotating object boxes. Second, as shown in Fig.~\ref{fig:top}, we design a QA framework to let the computer automatically figure out a limited number of typical examples of \textit{``known unknowns''} in the unannotated images, ask users questions, and use the answers to refine the AOG.

Thus, as shown in Fig.~\ref{fig:QAs}, we design six types of questions. Each question is oriented to a certain node in the AOG, \emph{e.g.} whether an image contains an object of a certain category and whether the current AOG provides a correct localization of an object (or a certain semantic part of a category). The computer uses these questions to overcome image noises caused by incorrect search results, intra-class variations, and ubiquitous occlusions.

Note that this multiple-shot QA learning does not fall within a conventional paradigm of active learning. First, we do not pre-define a certain feature space of an object as the prerequisite of active learning. Instead, we use the QA process to gradually enrich the knowledge of object structure, \emph{i.e.} discovering new alternative part appearance and mining the detailed components of each part. Second, we do not simply treat each answer as a single annotation of a specific object/part sample, but we generalize specific answers by mining the corresponding common patterns from big data in a weakly-supervised manner.

\textbf{Cost-sensitive policy:} We formulate a mixed loss for each node in the AOG as the unified paradigm to guide the learning of hierarchical object details. It includes a generative loss (measuring the model error in explaining the images) and a discriminative loss (\emph{i.e.} the model's fitness to human answers). Thus, among the six types of questions, each question corresponds to a certain node in the AOG, and we can use its answers to explicitly optimize the generative and/or discriminative loss of this node. Clear losses and semantic meanings of middle-layer nodes make our deep AOG different from deep neural networks.

As shown in Fig.~\ref{fig:top}, the QA framework uses the node loss to identify the nodes that are insufficiently trained, and selects the best sequence of questions to optimize a list of AOG nodes in an online manner. In each step, the QA framework balances the costs and potential gains of different questions, and selects the questions with high gains and low costs, to ensure high learning efficiency, which trades off the generative and discriminative losses, the human labor for annotations, and the computational cost.

In fact, this cost-sensitive policy is extensible. In this study, the QA framework combines six types of questions and four modules of 1) graph mining~\cite{OurICCV15AOG} (unsupervised mining of AOG structures without the labeling of object locations), 2) And-Or template learning~\cite{MiningAOG} (discovery of detailed structures within aligned parts), 3) supervised learning, and 4) object parsing. In addition, people can extend the QA system by adding more questions and modules.

\subsection{Related work}

\textbf{Knowledge organization for big data:} Many studies organized models of different categories in a single system. The CNN~\cite{CNN} encodes knowledge of thousands of categories in numerous neurons. The black-box representation of a CNN is not fully chaotic. \cite{InterpretabilitySurvey} made a survey of studies to understand feature representations in neural networks. For example, as shown in \cite{explanatoryGraph}, each filter in a convolutional layer usually encodes a mixture of visual concepts. For example, a filter may represent both the head part and the tail part of an animal. However, how to clearly disentangle different visual concepts from convolutional filters is still a significant challenge.

Recently, there has been a growing interest in modeling high-level knowledge beyond object detection. \cite{Gpt_NEIL,Gpt_KB} mined models for different categories/subcategories from web images. \cite{KB_Fei_Hierarchy} constructed a hierarchical taxonomic relationship between categories. \cite{T2V_1,T2V_2,VisualLanguageConcept,TextCNN} formulated the relationships between natural language and visual concepts. \cite{VisualQA} further built a Turing test system. \cite{Gpt_Context} modeled the contextual knowledge between objects. Knowledge in these studies was mainly defined upon object-level models (\emph{e.g.} the affordance and context). In contrast, we explore deep structures within objects. The deep hierarchy of parts provides a more informative understanding of object statuses.

\textbf{Multiple-shot QA for learning:} Many weakly-supervised methods and unsupervised methods have been developed to learn object-level models. For example, studies of \cite{WeaklyMIL,WeaklyDPM,WeaklyHOG,OurICCV15AOG}, object co-segmentation~\cite{Coseg2}, and object discovery~\cite{DiscoveryCNNFeature,MILDiscovery} learned with image-level annotations (without object bounding boxes). In particular, \cite{ChoDiscovery,ModelMining1} did not require any annotations during the learning process. \cite{WeblyConcept2,Gpt_WeaklyCNN,WeblyConcept3,WeblyConcept4} learned visual concepts from web images.

However, when we explore detailed object structures, manual annotations are still necessary to avoid model drift. Therefore, inspired by active learning methods~\cite{Active4,i13,Active1,Active2,Active3}, we hope to use a very limited number of human-computer QAs to learn each object pose/viewpoint. In fact, such QA ideas have been applied to object-level models~\cite{KB_Fei_Annotation,KB_Fei_InteractionLabel,TuQA}. Branson \emph{et al.}~\cite{ActivePart} used human-computer interactions to point out locations of object parts to learn part models, but they did not provide part boxes. In contrast, we focus on deep object structures. We design six types of human-computer dialogues/QAs for annotations (see Fig.~\ref{fig:QAs}). Our QA system chooses questions based on the generative and discriminative losses of AOG nodes, thereby explicitly refining different AOG nodes. In experiments, our method achieved good performance when we only label parts on 3--5 objects for each pose/viewpoint. Similarly, \cite{DeepQA} used active QA to learn a semantic tree to disentangle neural activations inside neural networks into hierarchical representations of object parts.

\textbf{Transparent representation of structures} is closely related to the deep understanding of object statuses. Beyond the object bounding box, we can further parse the object and align visual concepts at different layers to different object parts/sub-parts, which provides rich information of local appearance, poses, and viewpoints. In previous studies, many part models were designed with single-layer latent parts~\cite{DiscoveryCNNFeature,latentPart} or single-layer semantic parts~\cite{SSDPM,PCPChen,ActivePartModel,SingleImgPopup,RCNNDetailPart,ActivePart}, and trained for object detection with strong supervision. \cite{WeblyConcept3,WeblyConcept4} proposed to automatically learn multi-layer structures of objects from web images, which models the object identity, object viewpoints, semantic parts and their deformation locations. Whereas, we have a different objective, \emph{i.e.} weakly-supervised mining a nine-layer deep structural hierarchy of objects, which models detailed shape primitives of objects. \cite{interpretableCNN} learned an interpretable CNN with middle-layer filters representing object parts, and \cite{explanatoryTree_arXiv} further used an explanatory tree to represent the CNN's logic of using parts for object classification. \cite{explainer} learned an explainer network to interpret the knowledge of object parts encoded in a pre-trained CNN. \cite{transplant} further designed an interpretable modular structure for a neural network for multiple categories and multiple tasks, where each network module is functionally interpretable.


\subsection{Contributions}

The paper makes the following contributions:

\noindent
1) We propose a nine-layer AOG to represent the deep semantic hierarchy of objects.

\noindent
2) We propose an efficient QA framework that allows the computer to discover something unknown, to ask questions, and to explicitly learn deep object structures from human-computer dialogues.

\noindent
3) We use a general and extensible cost-sensitive policy to implement the QA system, which ensures a high efficiency of mining knowledge. To the best of our knowledge, our method is the first to reduce the cost of learning part localization to about ten annotations for each part.

\noindent
4) We can use our QA framework to learn deep semantic hierarchies of different categories from web images.

\begin{figure}
\centering
\includegraphics[width=\linewidth]{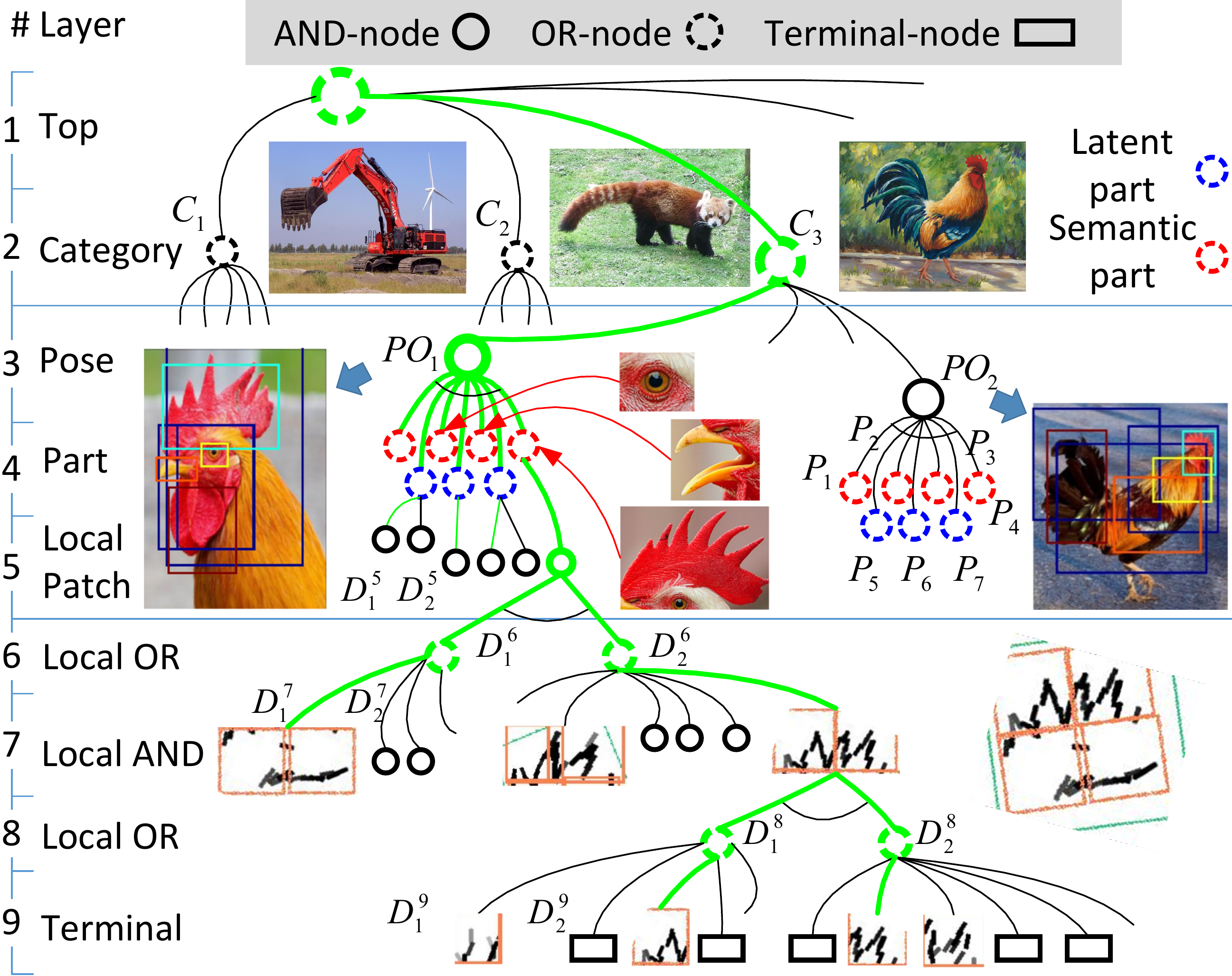}
\caption{A nine-layer And-Or Graph. An object can be explained by a \textit{parse graph} $\hat{pg}$, which is indicated by green lines. In the parse graph, AND nodes encode deformable structures between local patches, and OR nodes contain alternative local patterns. Each \textit{pose/viewpoint} has both \textit{latent parts} without names (blue OR nodes$\rightarrow$blue rectangles on roosters) and \textit{semantic parts} with specific names (red OR nodes$\rightarrow$other rectangles on roosters).}
\label{fig:AOG}
\end{figure}

\section{And-Or graph representation}

\begin{figure}
\centering
\includegraphics[width=0.9\linewidth]{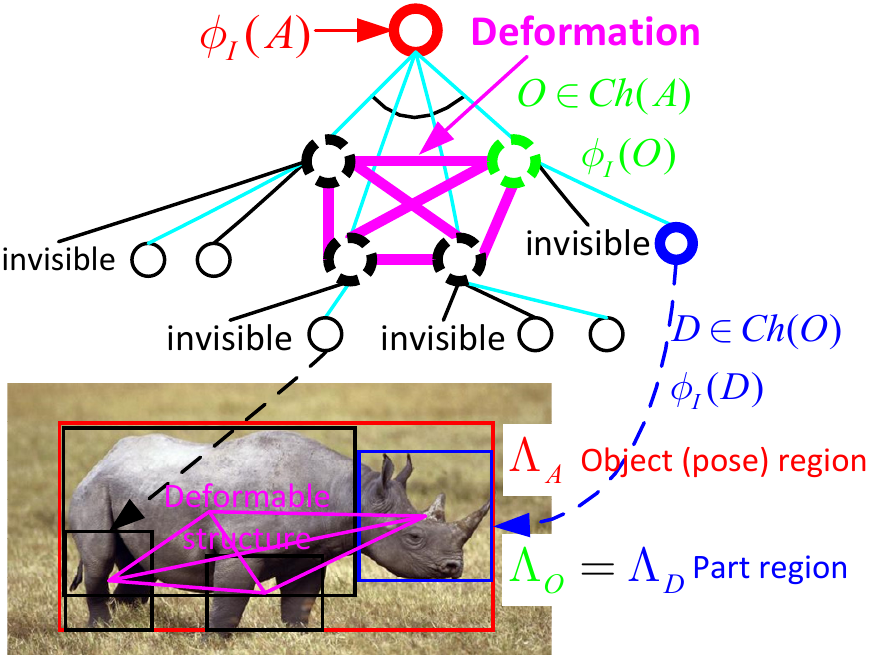}
\caption{Three-layer And-Or-And/Terminal structure in the AOG. Cyan lines indicate a parse graph for object inference.}
\label{fig:3Layer}
\end{figure}

Fig.~\ref{fig:AOG} shows the nine-layer AOG, which encodes visual concepts at different levels within objects and organizes their hierarchy. The basic element of the AOG is the three-layer And-Or structure in Fig.~\ref{fig:3Layer}, where an AND node represents 1) part compositions of a certain concept and 2) their deformation information, and an OR node lists alternative local patterns for a certain part. Let ${\boldsymbol\theta}$ denote all the AOG parameters. Let us use the AOG for object parsing in image $I$. For each node $D$ in the AOG, we use $\Lambda_{D}$ and ${\boldsymbol\theta}_{D}\!\subset\!{\boldsymbol\theta}$ to denote the image region corresponding to $D$ and the parameters related to $D$, respectively.

\textbf{Each Terminal node} $T$ in the bottom layer represents a pattern of local shape primitives. The reference score of node $T$ in image $I$ is formulated as
\begin{equation}
S_{I}(T)=\langle\omega_{T},\Phi(I_{\Lambda_{T}})\rangle
\label{eqn:OR}
\end{equation}
where $\Phi(I_{\Lambda_{T}})$ denotes the local features for the region $\Lambda_{T}$ in $I$, and ${\boldsymbol\theta}_{T}\!=\!\omega_{T}$ is the parameter.

\textbf{Each OR node} $O$ in the AOG provides a list of alternative local appearance patterns. In particular, OR nodes in Layers 1 and 2 encode the category choices and possible object poses/viewpoints within each category, respectively, and those in Layers 4, 6, and 8 offer local pattern candidates. When we use the AOG for object inference in image $I$, $O$ selects its child node with the highest score as the true configuration:
\begin{equation}
S_{I}(O)={\max}_{D\in Ch(O)}S_{I}(D)\\
\label{eqn:OR}
\end{equation}
where function $Ch(\cdot)$ indicates the children set of a node. The child node $D$ can be a Terminal node, an OR node, or an AND node. Note that $\textrm{``invisible''}\!\in\!Ch(O)$ is also a child of $O$, which is activated when other children patterns cannot be detected.

\textbf{Each AND node} $A$ in the AOG contains some sub-region components, and it models their geometric relationships. In particular, the AND nodes in Layer 3 organize the relationship between object \textit{poses/viewpoints} and object \textit{parts}, and those in Layers 5 and 7 encode detailed structural deformation within part patches. The inference score of $A$ is formulated as the sum of its children's scores:
\begin{equation}
\begin{split}
S_{I}(A)=&w_{A}\Big[S_{I}^{app}(A)+{\sum}_{D\in Ch(A)}S_{I}(D)\\
&+{\sum}_{(D,D')\in{\mathcal N}(A)}w_{DD'}S_{I}(D,D')\Big]+b_{A}
\end{split}
\label{eqn:AND}
\end{equation}
where {$S_{I}^{app}(A)$} represents the score of the global appearance in the region $\Lambda_{A}$. {${\mathcal N}(A)$} denotes the set of $A$'s neighboring children pairs. {$S_{I}(D,D')$} measures the deformation between image regions
$\Lambda_{D}$ and $\Lambda_{D'}$ of sibling children $D$ and $D'$. $w_{DD'}$ and $w_{A},b_{A}\!\in\!{\boldsymbol\theta}_{A}$ are constant weighting parameters for normalization. $w_{A}$ and $b_{A}$ are learned to make $S_{I}(A)$ have zero mean and unit variance through random background images.

\begin{figure*}
\centering
\includegraphics[width=\linewidth]{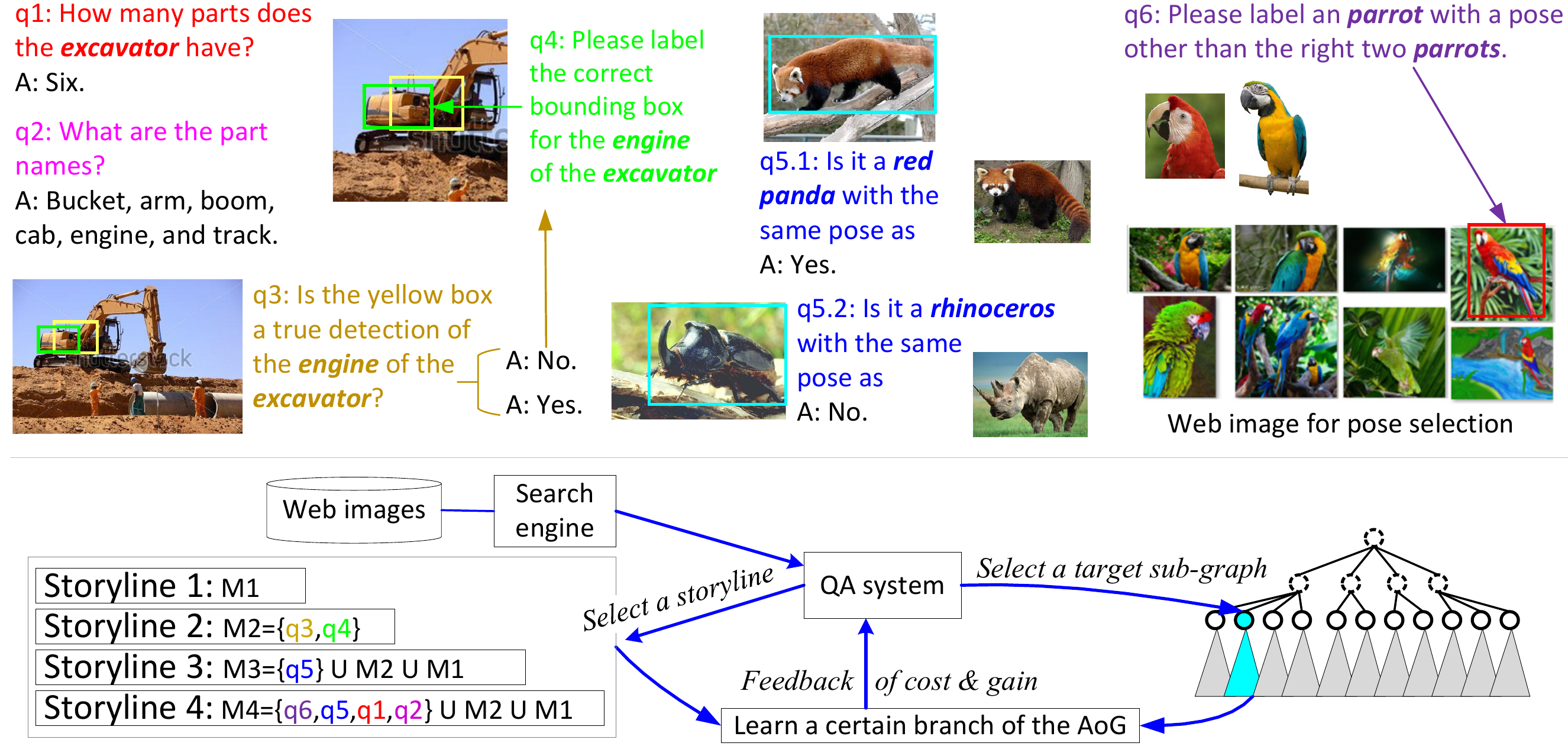}
\caption{Design of questions (top) and the QA framework (bottom). The QA framework iteratively selects a storyline and applies it to a target sub-AOG.}
\label{fig:QAs}
\end{figure*}

\subsection{Design of Layers 3--5}

\textbf{Layers ${\bf3\!\rightarrow\!4}$:} The three-layer And-Or structure that ranges across the \textit{pose/viewpoint}, \textit{part}, and \textit{local} layers is derived from the AOG pattern proposed in \cite{OurICCV15AOG}. This technique models the three-layer sub-AOG as the common subgraph pattern that frequently appears among a set of large graphs (\emph{i.e.} images). For each \textit{pose/viewpoint} node $PO$ under a category node $C$, we do not model its global appearance $S_{I}^{app}(PO)$. $PO$ contains two types of children nodes, \emph{i.e.} \textit{latent} children (the parts mined automatically from big data without clear semantic meaning) and \textit{semantic} children (the parts with certain names). Thus, based on (\ref{eqn:AND}), we can write the inference score of $PO$ as
\begin{small}
\begin{equation}
\!S_{I}(PO)\!=\!w_{PO}\Big[\!\!\!\!\!\!\sum_{D\in Ch(PO)}\!\!\!\!\!\!S_{I}(P)+\!\!\!\!\!\!\!\!\!\!\!\!\sum_{(P,P')\in{\mathcal N}(PO)}\!\!\!\!\!\!\!\!\!\!\!\!w_{PP'}S_{I}(P,P')\Big]+b_{PO}\!
\end{equation}
\end{small}

\textbf{Part deformation:} We connect all pairs of \textit{part} nodes under the pose/viewpoint $PO$ as neighbors. For each pair of \textit{part} nodes $(P,P')\!\in\!{\mathcal N}(PO)$, the deformation score between them measures the squared difference between the ideal (average) geometric relationship $\Phi(P,P')$ and the actual part relationship detected in the image $\Phi(\Lambda_{P},\Lambda_{P'})$. In addition, we also assign a specific deformation penalty $\rho$ as the deformation score, when one of the parts are not detected. The average geometric relationship $\Phi(P,P')$ and the penalty $\rho$ is estimated.
\begin{small}
\begin{equation}
\!S_{I}(P,P')=\left\{\begin{array}{ll}\rho, & \!\!\!\!\!\!\!\!\!\!\!\!\!\!\!\!\!\!\!\!\!\!\!\!\!\!\!\!\!\!\textrm{$P$ or $P'$ is not detected.}\\
+\infty, & \!\!\!\!\!\!\!\!\!\!\Lambda_{P}=\Lambda_{P'}\!\!\!\!\!\\
\Vert \Phi(P,P')-\Phi(\Lambda_{P},\Lambda_{P'})\Vert^2, & \!\!\textrm{otherwise}\end{array}\right.\!\!\!\!\!\!\!\!\!\!\!
\end{equation}
\end{small}
where the geometric relationships $\Phi(\Lambda_{P},\Lambda_{P'})$ between $P$ and $P'$ comprise three types of pairwise features, \emph{i.e.} 1) $\log(\frac{s_{P}}{s_{P'}})$, 2) $\frac{p_{P}-p_{P'}}{\Vert p_{P}-p_{P'}\Vert}$, and 3) $\log\frac{[s_{P},s_{P'}]}{\Vert p_{P}-p_{P'}\Vert}$. $s_{P}$ and $p_{P}$ denote the scale and 2D position of the part $P$, respectively.

\textbf{Layers ${\bf4\!\rightarrow\!5}$:} To simplify the AOG, we allow latent part nodes to have multiple children, but the semantic part node can only have one child besides the \textit{``invisible''} child. For each child $D$ in Layer 5 of a \textit{latent part}, its appearance score measures the squared difference between $D$'s ideal (average) appearance $\Phi(D)$ and the actual appearance detected in the image $\Phi(I_{\Lambda_{D}})$. Then, for the only child $D'$ of a \textit{semantic part}, we use part annotations to train a linear SVM to classify its local appearance, and set the appearance score of $D'$ as the SVM score. We also assign a specific appearance penalty $\rho_{D}$ for ``invisible'' children in Layer 5.
\begin{small}
\begin{equation}
\!S_{I}(D)\!=\!\left\{\!\!\begin{array}{ll}
w_{D}\Vert\Phi(D)-\Phi(I_{\Lambda_{D}})\Vert^2+b_{D}, & \!\!\!\textrm{$D$ is a latent part}\!\!\\
w_{D}SVM(\Phi(I_{\Lambda_{D}}))+b{D}, & \\
& \!\!\!\!\!\!\!\!\!\!\!\!\!\!\!\!\!\!\!\!\!\!\!\!\!\!\!\!\!\!\!\!\!\textrm{$D$ is a visible semantic part}\\
\rho_{D}, & \\
& \!\!\!\!\!\!\!\!\!\!\!\!\!\!\!\!\!\!\!\!\!\!\!\!\!\!\!\!\!\!\!\!\!\!\!\!\textrm{$D$ is an invisible semantic part}\!\!
\end{array}\right.\!\!\!\!\!\!\!
\end{equation}
\end{small}
where $w_{D}$ and $b_{D}$ are learned to make $S_{I}(D)$ have zero mean and unit variance through random background images. The appearance feature $\Phi(I_{\Lambda_{D}})$ for patch $D$ comprise the HOG features and the height-width ratio of the patch. A linear SVM is learned to estimate the score for a visible semantic part, which returns a positive/negative value if $I_{\Lambda_{D}}$ is a true/false detection of $D$. Model parameters, including average part appearance $\Phi(D)$, SVM parameters for semantic parts, the appearance penalty $\rho_{D}$ would be learned.

\subsection{Design of Layers 5--9}

The bottom four layers (Layers 6--9) of the AOG represent detailed structures within the \textit{semantic} patches in Layer 5 based on the And-Or template proposed in \cite{MiningAOG}. First, for each AND node $A$ in Layers 5 and 7, we do not encode its global appearance. $A$ has two children, and the deformation relationship between the two children is used to roughly model the ``\textit{geometric OR} relationships'' involved in \cite{MiningAOG}. Second, each OR node $O$ in Layers 6 and 8 has several children, which encodes only the ``\textit{structural OR} information'' described in \cite{MiningAOG}. Finally, terminal nodes in Layer 9 are described by the HIT feature mined by~\cite{LearnHIT}, which combines information of sketches, texture, flat area, and colors of a local patch.

\subsection{Object parsing (inference)}
\label{sec:inference}

Given an image $I$, we use the AOG to perform hierarchical parsing for the object inside $I$, \emph{i.e.} estimating a \textit{parse graph} (see green lines in Figs.\ref{fig:AOG}) to explain the object:
\begin{equation}
\hat{pg}={\arg\!\max}_{pg}S_{I}(pg)
\end{equation}
where we define the parse graph as a set of activated node regions for object understanding, {$\hat{pg}\!=\!\{\Lambda_{\hat{C}},\Lambda_{\hat{PO}},$ $\hat{\Lambda}_{P_{1}}, \hat{\Lambda}_{P_{2}}, \ldots, \hat{\Lambda}_{D^{9}_{1}}, \ldots, \hat{\Lambda}_{D^{9}_{n}}\}$}, which describes an inference tree of the AOG. We can understand the parse graph in a top-down manner. 1) Let an OR node $O$ in Layers 1, 2, 4, 6, or 8 have been activated and put into the parse graph ($\hat{\Lambda}_{O}\!\in\!\hat{pg}$). $O$ activates its best child {$\hat{D}={\arg\!\max}_{D\!\in\!Ch(O)}S_{I}(D)$} to explain the $O$'s image region $\Lambda_{\hat{D}}=\hat{\Lambda}_{O}$, and puts $\hat{D}$ into the parse graph ($\Lambda_{\hat{D}}\!\in\!\hat{pg}$). 2) Let an AND node $\hat{A}$ in Layers 3, 5, or 7 haven been activated and put into the parse graph ($\Lambda_{\hat{A}}\!\in\!\hat{pg}$). $\hat{A}$ determines the best image region inside $\Lambda_{\hat{A}}$ for each of its OR children $O\!\in\!Ch(\hat{A})$, \emph{i.e.} {$\{\hat{\Lambda}_{O}\}={\arg\!\max}_{\{\Lambda_{O}\}} S_{I}(A)|_{\{\Lambda_{O}\}}$}, and put $\{\hat{\Lambda}_{O}\}$ into the parse graph. Therefore, because we do not encode the global appearance of pose/viewpoint nodes, the objective of object parsing can be re-written as
\begin{small}
\begin{equation}
\begin{split}
\max_{pg}S_{I}(pg)&=\max_{PO\in\Omega_{pose}}S_{I}(PO)\\
&\!\!\!\!\!\!\!=\max_{PO\in\Omega_{pose}}\max_{\{\Lambda_{P}\}}w_{PO}\Big\{{\sum}_{P\in Ch(PO)}S_{I}(P)\\
&+{\sum}_{(P,P')\in{\mathcal N}(PO)}w_{PP'}S_{I}(P,P')+b_{PO}\Big\}
\end{split}
\end{equation}
\end{small}
where $\Omega_{pose}$ is the set of pose/viewpoint nodes in the AOG. The target parse graph $\hat{pg}$ for Layers 3--5 can be estimated via graph matching~\cite{OurICCV15AOG}. As mentioned in \cite{OurICCV15AOG}, (\ref{eqn:AND}) is a typical quadric assignment problem that can be directly solved by optimizing a Markov random field~\cite{TRWS}. The detailed inference for Layers 6--9 is solved by using \cite{MiningAOG}. The left-right symmetry of objects is considered in applications.

\begin{table*}[t]
\caption{Four types of storylines for each pose/viewpoint $PO_{i}$.}
\label{tab:storyline}
\begin{center}
\resizebox{1.0\hsize}{!}{\begin{tabular}{c|l|c|cccl}
\hline
\# & \multicolumn{1}{|c|}{Question stories $M_{i}$ for pose/viewpoint $PO_{i}$} & Participants $U_{i}$ & \!\!$\Delta L_{PO_{i}}^{gen}$\!\! & \!\!$\Delta L_{PO_{i}}^{cate}$\!\! & \!\!$\Delta L_{PO_{i}}^{part}$\!\! & \multicolumn{1}{c}{$Cost(Q_{i})$}\\
\hline
1 & retrain category classification & Computer & & \checkmark & & $C_{PO_{i}}^{ret}$\\
\hline
2 & check \& correct inaccurate semantic part localizations & Users & & \checkmark & \checkmark & $C_{PO_{i}}^{ckp}+C_{PO_{i}}^{lbp}$\\
\hline
3 & 1) QA-based collection of object samples for pose/viewpoint & Users & \checkmark & \checkmark & \checkmark & $C_{PO_{i}}^{col}+C_{PO_{i}}^{cko}+C_{PO_{i}}^{ckp}$\\
& $PO_{i}$, 2) mine the latent structure of pose/viewpoint $PO_{i}$ & \& Computer & & & &$+C_{PO_{i}}^{lbp}+C_{PO_{i}}^{ret}$\\
\hline
4 & generate a new pose/viewpoint: label an initial object example, & Instructors & \checkmark & & & $C_{PO_{i}}^{pose}+3C_{PO_{i}}^{col}+3C_{PO_{i}}^{cko}$\\
& collect samples, mine latent structure, label parts & \& Computer & & & & $+C_{PO_{i}}^{lbp}+C_{PO_{i}}^{ret}$\\
\hline
\end{tabular}}
\end{center}
\end{table*}

\section{Cost-sensitive QA-based active learning}

\subsection{Brief overview of QA-based learning}

In this section, we define the overall risk of the AOG. We use this risk to guide the growth of the AOG, which includes the selection of the questions, refining the current visual concepts in the AOG based on the answers, and mining new concepts as new AOG branches. The overall risk combines both the cost of asking questions during the learning process and the loss of AOG representation. The loss of AOG representation further comprises the generative loss (\emph{i.e.} the fitness between the AOG and real images) and the discriminative loss (\emph{i.e.} the AOG fitness to human supervision).

Therefore, the minimization of the AOG risk is actually to select a limited number of questions that can potentially minimize the AOG loss. In fact, we organize the six types of questions into four types of QA storylines (Fig.~\ref{fig:QAs}). In each step of the QA process, we conduct a certain storyline to decrease the risk. Meanwhile, we evaluate the gain (loss decrease) of different AOG nodes after each storyline, so that we can determine the next best storyline in an online manner.

Unlike previous active learning methods that directly use human annotations as ground-truth samples for training, we generalize specific annotations to common patterns among big data so as to update the AOG.

For example, in Layer 4 of the AOG, there are two types of \textit{parts}, \emph{i.e.} the \textit{semantic parts} and \textit{latent parts}. In Storylines 3 and 4 (details will be discussed later), we first 1) ask for object samples with a certain pose/viewpoint, 2) based on the object examples, select a large number of similar objects from all the web images as potential positives of this pose/viewpoint, then 3) use \cite{OurICCV15AOG} to mine the common part patterns among these objects as the \textit{latent parts}, and 4) model their spatial relationships.

Thus, as in (\ref{eqn:AND}) and Fig.~\ref{fig:3Layer}, spatial relationships between latent parts constitute a graph that represents the latent structure of the pose/viewpoint. Then, we continue to ask for semantic parts in Storylines 3 and 4, and use the pre-mined latent pose/viewpoint structure to localize relative positions of the newly annotated semantic parts. Such a combination of structure mining from big data and part annotations on small data ensures high learning stability.

In the following subsections, we introduce the detailed implementations of the proposed QA framework.

\subsection{Notation}

As shown in Fig.~\ref{fig:QAs}, we design six types of questions to learn the AOG, and organize these questions into four types of storylines. Let us assume that the QA framework has selected a sequence of storylines ${\bf Q}\!=\!\{Q_1,Q_2,\ldots\}$, and modified the AOG parameters to $\hat{\boldsymbol\theta}({{\bf Q}})$. We use the system \textit{risk}, $Risk({\bf Q})$, to evaluate the overall quality of the current status of QA-based learning. The objective of the QA framework is to select the storylines ${\bf Q}$ that can greatest decrease the overall \textit{risk}:
\begin{equation}
\begin{split}
\hat{\bf Q}=&{\arg\!\min}_{{\bf Q}}Risk({\bf Q})\\
Risk({\bf Q})=&{\bf L}(\hat{\boldsymbol\theta}({{\bf Q}}))+Cost({\bf Q})
\end{split}
\label{eqn:QA}
\end{equation}
The system risk comprises the cost of the storylines $Cost({\bf Q})$ and the loss (inaccuracy) of the current AOG ${\bf L}(\hat{\boldsymbol\theta}({{\bf Q}}))$. Thus, we can expect the QA system to select \textit{cheap} storylines $\hat{\bf Q}$ that greatly improve the model quality.

\textbf{Definition of ${{\bf Q}}$ and its cost:} Let $\Omega$ denote the set of storylines. Theoretically, there are four different storylines in each pose/viewpoint node in the AOG, which will be introduced later. The QA system selects a sequence of storylines ${\bf Q}=\{Q_{i}\in\Omega\}_{i=1,2,\ldots}$ to modify the AOG. Each storyline line $Q_{i}\!\in\!\Omega$ comprises a list of questions and learning modules. As shown in Table~\ref{tab:storyline}, we can represent the storyline as a three-tuple $Q_{i}=(M_{i},U_{i},PO_{i})$. $Q_{i}$ proposes some questions $M_{i}\!\subset\!\{q_1,q_2,\ldots,q_6\}$ ($q_{j}$ is a question defined in Fig.~\ref{fig:QAs}) for the target parse graph of the pose/viewpoint $PO_{i}$, expects a tutor $U_{i}$ to answer these questions, and then uses the answers for training. These storylines choose ordinary users, professional instructors, or the computer itself as the tutor $U_{i}$ to answer these queries. Because there are four types of storylines for each pose, the entire search space for storylines is given as $\Omega=\{Q_{i}|PO_{i}\in\Omega_{pose},M_{i}\in\{\textrm{Storyline-1},\ldots,\textrm{Storyline-4}\}\}$.

In addition, each storyline $Q_{i}$ has a certain cost $Cost(Q_{i})$ according to both the human labor of answering and the computational cost of model learning\footnote[2]{Professional instructors have higher labor cost considering their professional levels.}. The overall cost of ${\bf Q}$ is given as
\begin{equation}
Cost({\bf Q})={\sum}_{i}Cost(Q_{i})
\end{equation}

\textbf{Definition of the AOG loss:} Let ${\bf I}=\{I_1,I_2,\ldots\}$ be a comprehensive web image dataset governed by the underlying distribution $f(I)$. When we use our AOG (with parameters $\hat{\boldsymbol\theta}$) to explain the images in ${\bf I}$, we can formulate the overall loss as
\begin{equation}
{\bf L}(\hat{\boldsymbol\theta})=E_{I\sim f(I)}\Big[\underbrace{-S_{I}(pg^{*})}_{\textrm{\normalsize generative loss}}+\underbrace{L(pg^{*},\hat{pg}|\hat{\boldsymbol\theta})}_{\textrm{\normalsize discriminative loss}}\Big]
\label{eqn:loss}
\end{equation}
where $pg^{*}$ and $\hat{pg}$ indicate the true parse graph configuration and the estimated configuration of $I$, respectively. The generative loss measures the fitness between the image $I$ and its true parse graph $pg^{*}$, and the discriminative loss evaluates the classification performance.

The generative loss can be rewritten as
\begin{equation}
\begin{split}
E_{f}[-S_{I}(pg^{*})]=&{\sum}_{PO}P(PO)L_{PO}^{gen},\\
L_{PO}^{gen}=&E_{I\in{\bf I}_{PO}}[-S_{I}(PO)]
\end{split}
\label{eqn:gen_pose}
\end{equation}
where ${\bf I}_{PO}\subset{\bf I}$ represents a subset of images that contain objects belonging to the pose/viewpoint $PO$, and $L_{PO}^{gen}$ denotes the average generative loss of images in ${\bf I}_{PO}$. $\Lambda_{PO}\in pg^{*}$ indicates the true pose/viewpoint of the object inside $I$. $P(PO)=\vert{\bf I}_{PO}\vert/\vert{\bf I}\vert$ measures the probability of $PO$.

The discriminative loss for the pose/viewpoint $PO$ comprises the loss for category (pose/viewpoint) classification $L_{PO}^{cate}$ and the loss for part localization $L_{PO}^{part}$:
\begin{equation}
E_{f}\big[L({pg}^{*},\hat{pg}|\hat{\boldsymbol\theta})\big]={\sum}_{PO}P(PO)\big\{L_{PO}^{cate}+L_{PO}^{part}\big\}\\
\label{eqn:dis_pose}
\end{equation}
where {$L_{PO}^{cate}=V^{cate}E_{I\in{\bf I}_{PO}}\big\{\max\{0,\Delta(\hat{C},C^{*})\!+\!$ $[S_{I}(\hat{C})$ $-S_{I}(C^{*})]\}\big\}$, $L_{PO}^{part}\!=\!V^{part}$ $E_{I\in{\bf I}_{PO},P\in Ch(PO)}$ $\big\{\max\{0,$ $\Delta(\hat{\Lambda}_{P}$ $,\Lambda_{P}^{*})$ $+\![S_{I}(P)|_{\hat{\Lambda_{P}}}$ $-S_{I}(P)|_{\Lambda_{P}^{*}}]\}\big\}$}. {$\Lambda_{\hat{C}},$ $\hat{\Lambda}_{P}\!\in\!\hat{pg}$, $\Lambda_{C^{*}},\Lambda_{P}^{*}\in pg^{*}$}. $V^{cate}$ and $V^{part}$ represent prior weights for category classification and part localization, respectively (here, we set {$V^{cate}\!=\!1.0$, $V^{part}\!=\!1.0$}).

\begin{algorithm}[!htb]
\caption{Pseudo-code for the learning process}
\label{alg:pseudo}
\begin{algorithmic}
\STATE {\bf Input:} 1. Web images searched for $K$ categories
\STATE {\bf \verb|     |} 2. Iteration Number $N$
\STATE{{\bf Output:} AOG}
initialization\;
\FOR{k:=1 \TO K}
\STATE Ask $q_{1}$ and $q_{2}$
\STATE Apply Storyline 4 to the $k$-th category
\ENDFOR
\FOR{i:=1 \TO N}
\STATE Estimate $Q_{i}$ by determining $M_{i}$ and $PO_{i}\in\Omega_{pose}$.
\STATE {\bf Switch} {$M_{i}$} {\bf do}
\STATE {\bf\quad Case} {Storyline 1}
\STATE \qquad Mining hard negative samples
\STATE \qquad Retrain part classifiers
\STATE {\bf\quad Case} {Storyline 2}
\STATE \qquad Select samples of $PO_{i}$ without part annotations
\STATE \qquad Ask $q_{3}$ and $q_{4}$
\STATE \qquad Train part classifiers
\STATE \qquad Learn Layers 5–-9 via \cite{MiningAOG}
\STATE {\bf\quad Case} {Storyline 3}
\STATE \qquad Collect new samples for $PO_{i}$
\STATE \qquad Ask $q_{5.1}$ and $q_{5.2}$
\STATE \qquad Graph mining~{\small\cite{OurICCV15AOG}} to learn Layers 3--5
\STATE \qquad Apply Storyline 2
\STATE \qquad Apply Storyline 1
\STATE {\bf\quad Case} {Storyline 4}
\STATE \qquad Ask $q_{6}$ to obtain the new target $PO_{i}$
\STATE \qquad Apply Storyline 3
\ENDFOR
\end{algorithmic}
\end{algorithm}

\subsection{Learning procedure}

Algorithm~\ref{alg:pseudo} summarizes the procedure of the QA-based active learning. In the beginning, we construct the top two layers of the AOG to contain a total of $K$ categories. We use keywords of these categories to crawl web images of the $K$ categories from the internet, and build a comprehensive web image dataset ${\bf I}=\{I_1,I_2,\ldots\}$. Next, we apply Storyline 4 to each category, which mines an initial model for a certain pose/viewpoint of this category. Then, we simply use a greedy strategy to solve (\ref{eqn:QA}), which estimates an optimal sequence of storylines ${\bf Q}=\{Q_{i}\}_{i=1,2,\ldots}$. In each step $i$, we recursively determine the next best storyline, $\hat{Q}_{i}$, as follows.
\begin{equation}
\hat{Q}_{i}={\arg\!\max}_{Q_{i}\in\Omega}\frac{-\Delta{\bf L}(\hat{\boldsymbol\theta}({\bf Q}))}{Cost(Q_{i})}
\end{equation}
where $\Delta{\bf L}(\hat{\boldsymbol\theta}({\bf Q}))$ denotes the potential AOG gain (decrease of the AOG loss, which is estimated by historical operations and introduced later) from storyline ${\bf Q}$. Considering (\ref{eqn:gen_pose}) and (\ref{eqn:dis_pose}), we can rewrite the above equation as
\begin{small}
\begin{equation}
\hat{Q}_{i}=\underset{{Q}_{i}\!=\!(M_{i},U_{i},PO_{i})}{\arg\!\max}\!\frac{-P(PO_{i})[\Delta L_{PO_{i}}^{gen}\!+\!\Delta L_{PO_{i}}^{cate}\!+\!\Delta L_{PO_{i}}^{part}]}{Cost({Q}_{i})}\!
\label{eqn:pi}
\end{equation}
\end{small}
where $\Delta L_{PO_{i}}^{gen}$, $\Delta L_{PO_{i}}^{cate}$, and $\Delta L_{PO_{i}}^{part}$ are the potential gains of $L_{PO_{i}}^{gen}$, $L_{PO_{i}}^{cate}$, and $L_{PO_{i}}^{part}$ after storyline $\hat{Q}_{i}$, respectively. $P(PO_{i})$ can be estimated based on the current web images collected for $PO_{i}$ (\emph{i.e.} $\hat{\bf I}_{PO_{i}}$)\footnote[3]{$\hat{\bf I}_{PO_{i}}$ denotes the current images that are collected for pose/viewpoint $PO_{i}$ from a category's image pool ${\bf I}_{C}$ in Storyline 3.} and the yes/no answer ratio during sample collection in Storyline 3.

\subsection{Introduction of storylines}

\textbf{Storyline 1: retraining category classification.} As the QA framework collects more and more web images, in this storyline, we use these images to update the AOG parameters for the classification of a certain pose/viewpoint $PO_{i}$. This storyline mainly decreases the discriminative loss {$L_{PO_{i}}^{cate}$}.

Given all the web images that have been collected for pose/viewpoint $PO_{i}$ (\emph{i.e.} $\hat{\bf I}_{PO_{i}}$\textcolor{red}{\footnotemark[3]}) we use the current AOG for object inference on these images. Given an incorrect object inference (\emph{i.e.} an image is incorrectly recognized as a pose/viewpoint $PO_{j}$ other than the true pose/viewpoint $PO_{i}$), we can use this inference result to produce hard negatives of semantic object parts for $PO_{j}$, and retrain its part classifier in Layer 5.

Therefore, the potential cost for a future storyline $Cost(Q_{i})$ mainly comprises the computational cost of object inference {$C_{PO_{i}}^{ret}\!=\!\lambda^{ret}\vert\hat{\bf I}_{PO_{i}}\vert\vert\Theta_{pose}\vert$}, where $\Theta_{pose}$ is the set of all the pose/viewpoint nodes, and $\lambda^{ret}$ is a weighting parameter\footnote[4]{Please see Section~\ref{sec:detail} for parameter settings of $\lambda^{\textrm{``x''}}$.}. The potential gain $\Delta L_{PO_{i}}^{cate}$ can be predicted simply using historical gains from similar storylines for pose/viewpoint $PO_{i}$\footnote[5]{Among all the storylines {$Q_{j}$, $j\!=1,\ldots,i\!-\!1$} before {$Q_{i}$}, we select the storylines that have both the same type of questions $M_{j}\!=\!M_{i}$ and the same target pose/viewpoint $PO_{j}\!=\!PO_{i}$ as {$Q_{i}$}. We record gains of $\Delta L_{PO_{i}}^{cate}$ and $\Delta L_{PO_{i}}^{part}$ after these storylines, and use these historical gains to predict the gain for a further storyline $pi_{i}$.}.

\textbf{Storyline 2: checking\,\&\,labeling semantic parts.} In this storyline, the computer 1) selects a sequence of images, 2) asks users whether the current AOG can correctly localize the semantic parts in these images, and 3) lets users correct the incorrect part localizations to update the AOG.

First, the QA system uses the pose/viewpoint model of $PO_{i}$ for object inference on the images $\hat{\bf I}_{PO_{i}}\subset\hat{\bf I}_{PO_{i}}^{unlabeled}$ in which semantic parts are not labeled. Next, the QA system selects a set of images that potentially contain incorrect localizations of semantic parts. We select the object samples that have good localizations of latent parts but inaccurate localizations of semantic parts, \emph{i.e.} having high scores for latent parts but low scores for semantic parts. Thus, we can determine the target sample (image) as $\hat{I}\!=\!{\arg\!\max}_{I\!\in\!\hat{\bf I}_{PO_{i}}^{unlabeled}}S_{I}(PO_{i})-S_{I}(PO_{i}^{lat})$, where $PO_{i}^{lat}$ is a dummy pose/viewpoint that is constructed by eliminating semantic parts from the current pose/viewpoint.

Then, the computer asks users to check whether the part localizations on the selected images are correct or not\footnote[6]{The QA system asks about part compositions/names for pose/viewpoint $PO_{i}$ in the first time of part labeling (see Fig.~\ref{fig:QAs}($q_1,q_2$)).} (see Fig.~\ref{fig:QAs}($q_3$)), and finally asks users to label the boxes for the incorrect part localizations (see Fig.~\ref{fig:QAs}($q_4$)).

Given the annotations of semantic part boxes, we update the geometric relationships between \textit{part} nodes in Layer 4 based on \cite{OurICCV15AOG}, and update SVM classifiers for local patch appearance in Layer 5. Given the part annotations, we can further learn detailed structures in Layers 5--9 via \cite{MiningAOG}.

The cost $Cost(Q_{i})$ of this storyline mainly comprises the human labor required for both part checking $C_{PO_{i}}^{ckp}$ and part labeling $C_{PO_{i}}^{lbp}$, which can be measured as {$C_{PO_{i}}^{ckp}\!=\!\lambda^{ckp}\textcolor{red}{\footnotemark[4]}\vert{SemanticCh(PO_{i})}\vert$, and $C_{PO_{i}}^{lbp}\!=\!\lambda^{lbp}\textcolor{red}{\footnotemark[4]}\vert{SemanticCh(PO_{i})}\vert$}, respectively. This storyline mainly decreases $L_{PO_{i}}^{cate}$ and $L_{PO_{i}}^{part}$. The potential gain $\Delta L_{PO_{i}}^{cate}$ and $\Delta L_{PO_{i}}^{part}$ for a future storyline can be predicted using historical gains\textcolor{red}{\footnotemark[5]}.

\textbf{Storyline 3: collecting \& labeling new samples.} This storyline collects new sample for pose/viewpoint $PO_{i}$ from web images to update the pose/viewpoint. It decreases the generative loss $L_{PO_{i}}^{gen}$ and the pose/viewpoint classification loss {$L_{PO_{i}}^{cate}$}. First, we use the sub-AOG of pose/viewpoint $PO_{i}$ to collect new samples from web images\footnote[7]{The images collected from search engines comprise both correct images with target objects and irrelevant background images.} with top inference scores. The system collects {$N\!=\!3\!\cdot\!1.5^{k}\!$} new samples, when it is the $k$-th time to perform Storyline 3 to pose/viewpoint $PO_{i}$.

Second, we randomly select $n$ ({$n\!=\!10$, here}) new object samples, ask users whether they are true samples with pose/viewpoint $PO_{i}$, and expect yes/no answers (see Fig.~\ref{fig:QAs}($q_{5.1},q_{5.2}$)).

Third, given the true samples, we use graph mining~\cite{OurICCV15AOG} to refine the And-Or structure in Layers 3--5 for $PO_{i}$. The sub-AOG is refined towards the common subgraph pattern (pose/viewpoint model) embedded in a set of large graphs (images). Its objective can be roughly written as follows, which is proved in Appendix~\ref{sec:append}.
\begin{equation}
\underset{{\boldsymbol\theta}_{PO_{i}}}{\arg\!\max}\!\underset{I\in\hat{\bf I}_{PO_{i}}}{E}[S_{I}(PO_{i})]\exp[-{\textrm{ModelComplexity}}({\boldsymbol\theta}_{PO_{i}})]
\label{eqn:graphMining}
\end{equation}
The above equation refines the ${\boldsymbol\theta}_{PO_{i}}$ by 1) adding (or deleting) new (or redundant) latent parts $P\in LatentCh(PO_{i})$ from the pose/viewpoint $PO_{i}$, 2) determine the children number (\emph{i.e.} the number of patches in Layer 5) of each latent part $P$, 3) updating the average appearance $\Phi(D)$ of each patch $D\in Ch(P)$, and 4) refining the average geometric relationship $\Phi(P,P')$ between each pair of children parts $P,P'\in Ch(PO_{i})$.

At the end of Storyline 1, we further apply Storylines 2 and 1 to refine semantic parts for pose/viewpoint $PO_{i}$ and retrain for pose/viewpoint classification.

Therefore, the potential cost of a future storyline can be computed as {$Cost(Q_{i})\!=\!C_{PO_{i}}^{col}\!+\!C_{PO_{i}}^{cko}\!+\!C_{PO_{i}}^{ckp}\!+\!C_{PO_{i}}^{lbp}\!+\!C_{PO_{i}}^{ret}$}. {$C_{PO_{i}}^{col}\!=\!\lambda^{col}\textcolor{red}{\footnotemark[4]}\vert{\bf I}_{C}\vert$} is the computational cost of sample collection, where ${\bf I}_{C}$ denotes the entire web image pool of category $C$, {$PO_{i}\!\in\!Ch(C)$}. {$C_{PO_{i}}^{cko}\!=\!\lambda^{cko}\textcolor{red}{\footnotemark[4]}n$} indicates the human labor of checking samples. $C_{PO_{i}}^{ckp}$, $C_{PO_{i}}^{lbp}$, and $C_{PO_{i}}^{ret}$ denote the costs of checking parts, labeling parts, and retraining pose/viewpoint classification, respectively, and can be estimated as introduced in Storylines 1 and 2. This storyline mainly decreases $L_{PO_{i}}^{gen}$, $L_{PO_{i}}^{cate}$ and $L_{PO_{i}}^{part}$. For the term of $L_{PO_{i}}^{gen}$, we can roughly estimate $P(PO_{i})\Delta L_{PO_{i}}^{cate}$ as {$-\textrm{mean}_{I\!\in\!\hat{\bf I}_{C}}\Delta S_{I}(C)$, $PO_{i}\!\in\!C$} in the last Storyline 3. {$\Delta L_{PO_{i}}^{cate},\Delta L_{PO_{i}}^{part}$} are approximated using historical gains\textcolor{red}{\footnotemark[5]}.

\textbf{Storyline 4: labeling a new sibling pose/viewpoint.} As shown in Fig.~\ref{fig:QAs}($q_6$), in this storyline, the QA system requires a professional instructor to label an initial sample for a new pose/viewpoint $PO_{i}$ in category $C$, and uses iterative graph mining~\cite{OurICCV15AOG} to extract the structure of Layers 3--5 for pose/viewpoint $PO_{i}$ (only mining latent parts in Layer 5). The graph mining is conducted with three iterations. In each iteration, we first collect new object samples for pose/viewpoint $PO_{i}$, as shown in Fig.~\ref{fig:QAs}($q_{5.1},q_{5.2}$). Based on the collected samples, we optimize the mining objective in (\ref{eqn:graphMining}) to mine/refine the latent parts in Layer 4 and the patches in Layer 5 for this pose/viewpoint. In this way, we obtain the latent structure of the new pose/viewpoint $PO_{i}$, and then we apply Storylines 2 to pose/viewpoint $PO_{i}$ to ask and label semantic parts and to fix these semantic parts on this latent structure. Finally, we apply Storyline 1 to train classifiers of the semantic parts for pose/viewpoint classification.

\begin{figure*}[th]
\centering
\includegraphics[width=\linewidth]{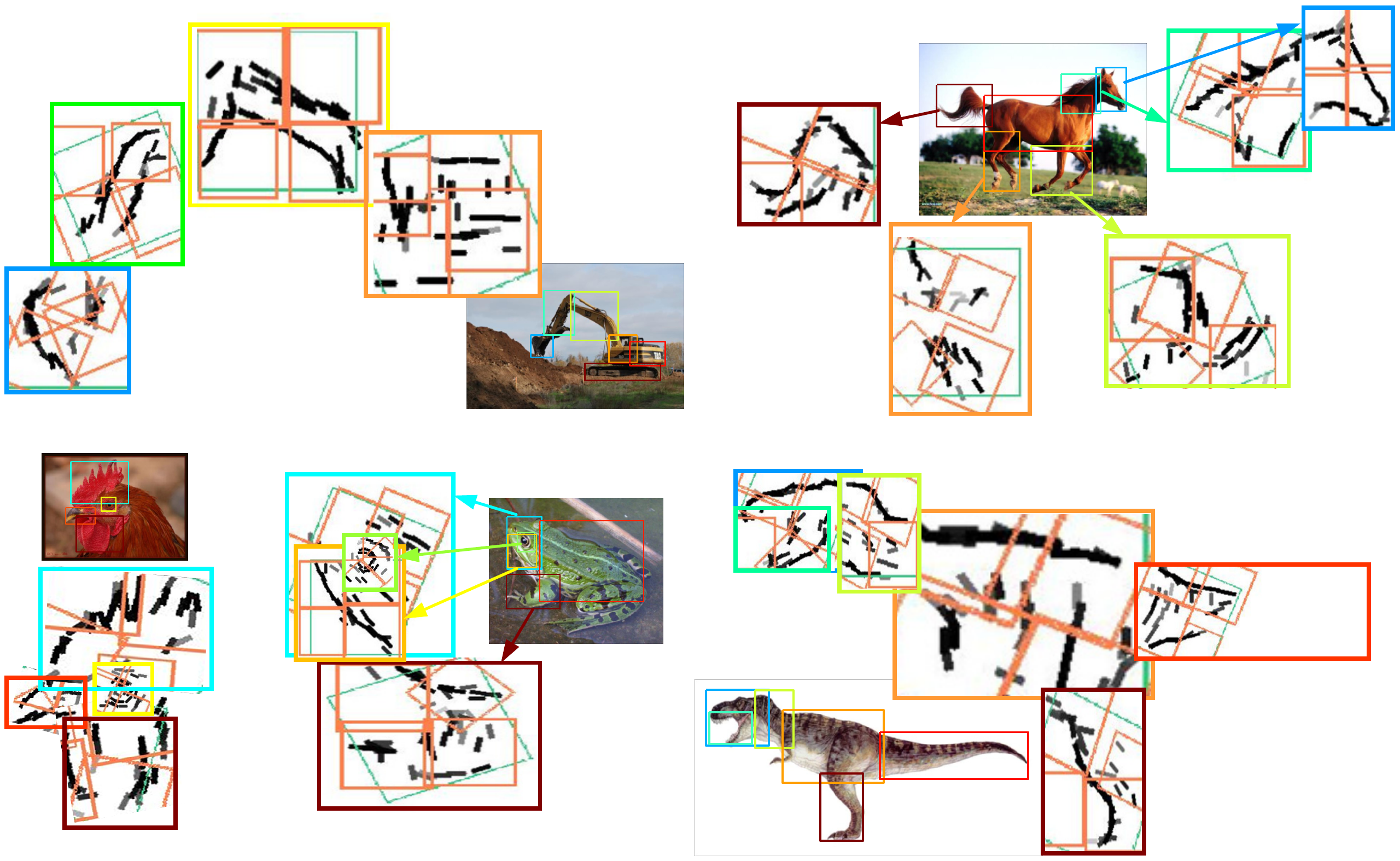}
\caption{Deep semantics within object parts. We mine the common structure within each object part, and represent the shape primitives in Layers 5--9 of the AOG. In fact, some of these shape primitives have certain latent semantics, \emph{e.g.} the mandible of a Tyrannosaurus rex within its ``mouth'' part. Given an image, the shape primitives can be aligned to their corresponding image regions with a certain deformation.}
\label{fig:result_details_1}
\end{figure*}

Therefore, the storyline cost is given as {$Cost(Q_{i})\!=\!C_{PO_{i}}^{pose}\!+\!3C_{PO_{i}}^{col}\!+\!3C_{PO_{i}}^{cko}\!+\!C_{PO_{i}}^{lbp}\!+\!C_{PO_{i}}^{ret}$}, where $C_{PO_{i}}^{pose}\!=\!\lambda^{pose}$ is a constant cost for labeling a new pose/viewpoint\textcolor{red}{\footnotemark[4]}, and other costs can be estimated as mentioned above. This storyline mainly decreases $L_{PO_{i}}^{gen}$, which can be computed as in Storyline 3.

\section{Experiments}

\subsection{Details}
\label{sec:detail}

To implement the QA system, we set the parameters as follows. $\lambda^{ckp}\!=\!1.0$, $\lambda^{cko}\!=\!1.0$, and $\lambda^{lbp}\!=\!5$. It is because that we found that the time cost of labeling a part is usually five times greater than that of making a yes/no judgment in our experiments. The computational cost of the collection/inference of each object was set as $\lambda^{ret}\!=\!0.01$, $\lambda^{col}\!=\!0.01$. We set $\lambda^{pose}\!=\!50$ as the labeling cost for a new pose/viewpoint.

We applied Bing Search and used 16 different keywords to collect web images. The keywords included ``\textit{bulldozer},'' ``\textit{crab},'' ``\textit{excavator},'' ``\textit{frog},'' ``\textit{parrot},'' ``\textit{red panda},''  ``\textit{rhinoceros},'' ``\textit{rooster},'' ``\textit{Tyrannosaurus rex},'' ``\textit{horse},'' ``\textit{equestrian},'' ``\textit{riding motorbike},'' ``\textit{bus},'' ``\textit{aeroplane},'' ``\textit{fighter jet},'' and ``\textit{riding bicycle}.'' With each keyword, we collected the top-1000 returned images. We used images of the first ten keywords to learn an AOG (namely \textit{AOG-10}) with ten categories to evaluate the learning efficiency of our QA framework. Then, we used images of the last seven keywords to learn an AOG (namely \textit{AOG-7}) with five categories (\textit{horse}, \textit{motorbike}, \textit{bus}, \textit{aeroplane}, and \textit{bicycle}) and tested the performance on the Pascal VOC dataset~\cite{PascalVOC}.

\subsection{Mining of the deep semantic hierarchy}

\begin{figure*}[th]
\centering
\includegraphics[width=\linewidth]{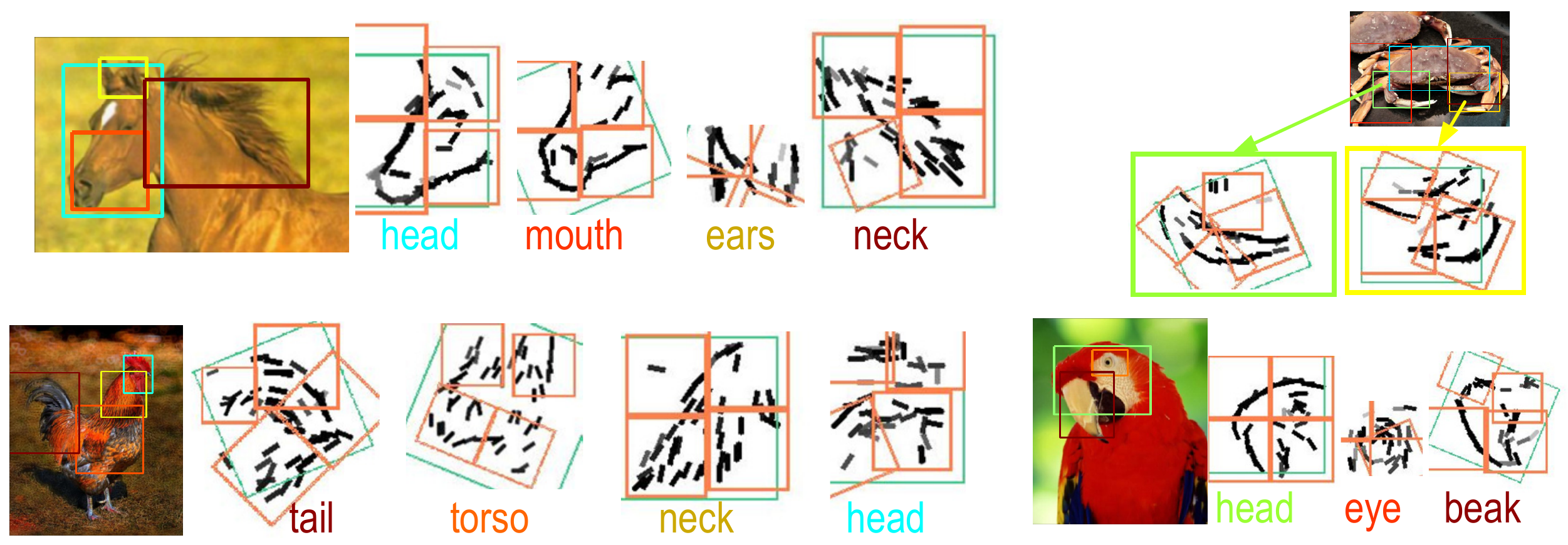}
\caption{Deep semantics within object parts. We mine the common structure within each object part, and represent the shape primitives in Layers 5--9 of the AOG.}
\label{fig:result_details_2}
\end{figure*}

Figs.~\ref{fig:result_details_1} and \ref{fig:result_details_2} illustrate the deep structures of some categories in the \textit{AOG-10}. The QA system applied a total of 39 storylines to learn \textit{AOG-10}. The \textit{AOG-10} contains two poses/viewpoints for the \textit{frog}, \textit{horse}, and \textit{parrot} categories, and three poses/viewpoints for each of the other seven categories in Layer 3. \textit{AOG-10} has 132 semantic part nodes and 84 latent part nodes in Layer 4. \textit{AOG-7} contains a total of 12 pose/viewpoint nodes in Layer 3, 48 semantic part nodes, and 48 latent part nodes in Layer 4.


\subsection{Evaluation of part localization}

The objective of this work is to learn a transparent representation of deep object hierarchy, and it is difficult to evaluate the quality of deep structures. Therefore, we evaluate our AOGs in terms of part localization, although our contribution is far more than it. We tested the \textit{AOG-10} on web images and tested the \textit{AOG-7} on the Pascal VOC dataset for a comprehensive evaluation.

\textbf{Baselines:} Our AOGs were learned with part annotations on only 2--14 objects in each category, but most previous methods require a large number of part annotations to produce a valid model. Nevertheless, we still selected the nine baselines for comparisons, including benchmark methods for object detection (here is part detection), popular part-localization approaches, and methods for interactive learning of parts. For each baseline, we randomly selected different numbers of training samples to learn the model and enable a fair comparison.

\begin{figure*}
\centering
\includegraphics[width=\linewidth]{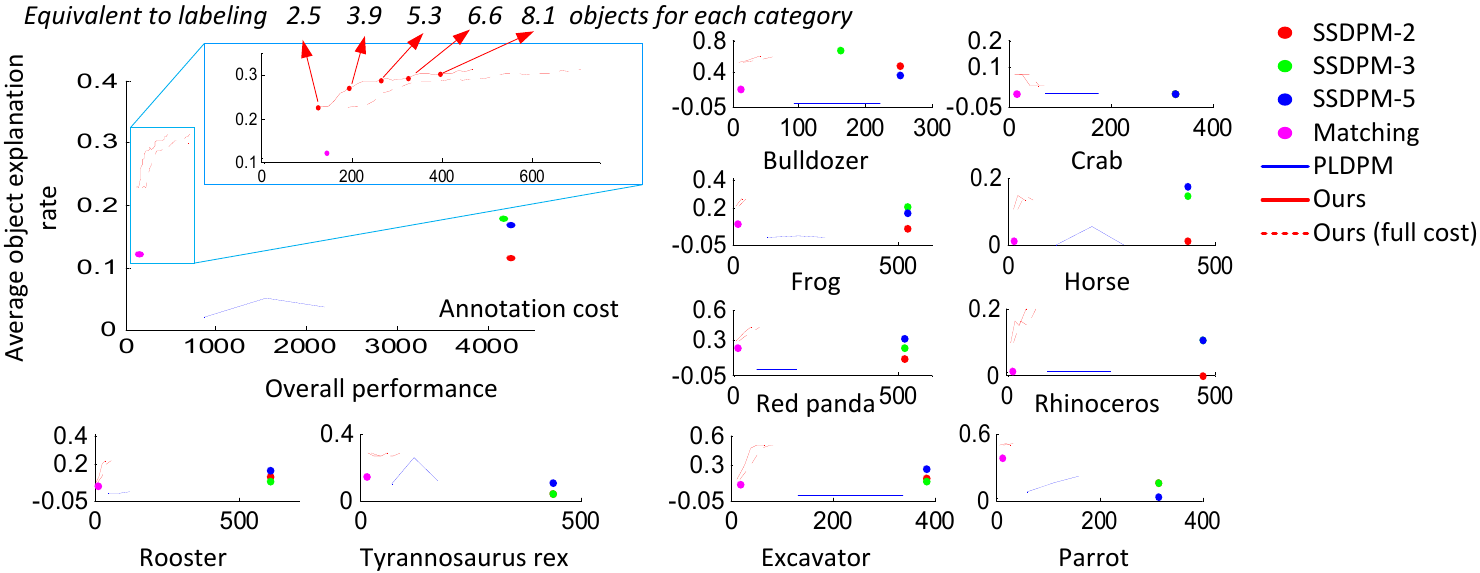}
\caption{Comparisons in the efficiency of knowledge mining. The annotation cost (horizontal axis) is computed based on \textbf{part} annotations. The top line shows such annotations are equivalent to labeling how many \textbf{objects} for each category. Instead of preparing a large training set for supervised methods, our method can achieve multiple-shot learning (on average, 2--10 shots for each part, here).}
\label{fig:compare_all}
\end{figure*}

First, we focused on \cite{SSDPM}, which uses annotations of semantic parts to train DPMs. This method clusters training samples to different object poses/viewpoints, and trains a DPM component for each pose/viewpoint. We designed three baselines based on \cite{SSDPM}, namely \textit{SSDPM-2}, \textit{SSDPM-3} and \textit{SSDPM-5}. For each category, \textit{SSDPM-2}, \textit{SSDPM-3} and \textit{SSDPM-5} learned two, three, and five pairs of left-right symmetric poses/viewpoints, respectively\footnote[8]{Due to the limited number of training samples, the \textit{bulldozer} and \textit{horse} categories could produce at most four pairs of pose/viewpoint models for \textit{SSDPM-5}. Training samples used in the baselines will be published after the paper acceptance.}.

Then, we used the technique of \cite{PLDPM} as the fourth baseline, namely \textit{PLDPM}, which required annotations of both the parts and object poses/viewpoints for training. To enable a fair comparison, we only collected and labeled training samples that corresponded to the poses/viewpoints in our AOG.

\begin{table*}[t]
\begin{center}
\caption{Performance of part localization}
\label{tab:VOC}
\resizebox{1.0\linewidth}{!}{\begin{tabular}{|c|ccccc|ccccc|ccccc|}
\hline
& & \multicolumn{2}{c}{bicyc-L} & \multicolumn{2}{c|}{bicyc-R} & & \multicolumn{2}{c}{bus-L} & \multicolumn{2}{c|}{bus-R} & & \multicolumn{2}{c}{aero-L} & \multicolumn{2}{c|}{aero-R}\\
& \#box & APP & AER & APP & AER & \#box & APP & AER & APP & AER & \#box & APP & AER & APP & AER\\
\hline
{ SSDPM}~\cite{SSDPM}
& 228 {\bf :}
& 58.7 & 58.4
& 67.2 & {\bf 65.7}
& 98 {\bf :}
& 30.0 & {\bf 36.8}
& 20.6 & 30.7
& 133 {\bf :}
& 13.9 & 22.2
& {\bf 24.6} & 31.0
\\
& 110 {\bf :}
& 53.2 & 55.0
& 54.7 & 58.5
&  57 {\bf :}
& 3.3 & 23.2
& 5.9 & 18.0
& 68 {\bf :}
& 7.7 & 15.4
& 12.3 & 31.8
\\\hline
{ P-Graph}~\cite{PCPChen}
& 204 {\bf :}
& 8.3 & 0
& 5.4 & 0
& 152 {\bf :}
& 11.2 & 0
& 15.9 & 0
& 156 {\bf :}
& 1.7 & 0
& 1.7 & 0
\\\hline
{ Fast-RCNN}~\cite{FastRCNN}
& 222 {\bf :}
& 23.0 & 2.6
& 21.3 & 1.8
& 109 {\bf :}
& 19.0 & 0
& 20.1 & 6.7
& 95 {\bf :}
& 14.6 & 4.2
& 16.8 & 2.3
\\
& 113 {\bf :}
& 24.1 & 5.1
& 15.1 & 0
& {\bf 51} {\bf :}
& 3.0 & 0
& 12.6 & 6.7
& 49 {\bf :}
& 5.7 & 0
& 7.0 & 0
\\\hline
YOLOv3~\cite{YOLOv3}
& 222 {\bf:}
& 33.8 & --
& 44.4 & --
& 109 {\bf:}
& 18.9 & --
& 12.3 & --
& 186 {\bf:}
& {\bf 14.9} & --
& 23.4 & --
\\\hline
{ Our}
& {\bf 9} {\bf :}
& {\bf 60.6} & {\bf 60.5}
& {\bf 68.8} & 65.1
& 54 {\bf :}
& {\bf 36.7} & 35.4
& {\bf 35.3} & {\bf 41.7}
& {\bf 24} {\bf :}
& 13.9 & {\bf 28.4}
& 17.5 & {\bf 31.0}
\\\hline\hline
& & \multicolumn{2}{c}{motor-L} & \multicolumn{2}{c|}{motor-R} & & \multicolumn{2}{c}{horse-L} & \multicolumn{2}{c|}{horse-R} & & & & &\\
& \#box & APP & AER & APP & AER & \#box & APP & AER & APP & AER & & & & &\\
\hline
{ SSDPM}~\cite{SSDPM}
& 30 {\bf :}
& 57.9 & 57.3
& 24.5 & 41.5
& 104 {\bf :}
& 10.1 & {\bf 40.6}
& 9.5 & 35.5
&
& &
& &
\\
& 24 {\bf :}
& 0 & 7.7
& 0 & 8.0
& 52 {\bf :}
& 0 & 18.7
& 1.4 & 16.2
&
& &
& &
\\\hline
{ P-Graph}~\cite{PCPChen}
& 148 {\bf :}
& 7.1 & 0
& 9.3 & 0
& 180 {\bf :}
& 3.7 & 0
& 0.6 & 0
&
& &
& &
\\\hline
{ Fast-RCNN}~\cite{FastRCNN}
& 163 {\bf :}
& 29.2 & 5.5
& 24.4 & 0
& 208 {\bf :}
& 29.7 & 8.3
& 26.1 & 8.5
&
& &
& &
\\
& 83 {\bf :}
& 15.6 & 1.8
& 9.7 & 0
& 104 {\bf :}
& 14.1 & 1.7
& 19.2 & 3.4
&
& &
& &
\\\hline
YOLOv3~\cite{YOLOv3}
& 163 {\bf:}
& 48.3 & --
& 30.6 & --
& 208 {\bf:}
& {\bf44.4} & --
& {\bf38.8} & --
&
& &
& &
\\\hline
{ Our}
& {\bf 9} {\bf :}
& {\bf 57.9} & {\bf 62.4}
& {\bf 32.7} & {\bf 48.6}
& {\bf 46} {\bf :}
& 24.6 & 35.8
& 23.0 & {\bf 35.7}
&
& &
& &
\\\hline
\end{tabular}}
\end{center}
\vspace{5pt}{\textit{\#box} indicates the number of \textbf{part} annotations for model learning, and the performance is evaluated by the values of (APP / AER). With the help of massive web images, our method only required $3\%$--$95\%$ number of the part annotations that were used by \textit{SSDPM}, and achieved comparable performance to \textit{SSDPM}.}
\end{table*}

The fifth baseline was another part model proposed by \cite{PCPChen}, namely \textit{P-Graph}, which organized object parts into a graph and trained an SVM based on the part appearance features and inter-part relationships for part localization.

The sixth baseline was image matching, namely \textit{Matching}, introduced in \cite{OurICCV15AOG}. Unlike conventional matching between automatically detected feature points~\cite{Cho1,LearnMatchingCMU_IJCV,LearnMatchingMaxMargin}, \textit{Matching} used a graph template to match semantic parts of objects in images. For a fair comparison, \textit{Matching} constructed a graph template for each pose/viewpoint in our AOG (\emph{i.e.} using the template of the initial sample labeled in Storyline 4).

Then, we used two benchmark methods for object detection, \emph{i.e.} \textit{Fast-RCNN}~\cite{FastRCNN} and YOLOv3~\cite{YOLOv3}, as the seventh and eighth baselines to detect object parts. For the fast-RCNN baseline, we chose the widely used 16-layer VGG network (VGG-16)~\cite{VGG} that was pre-trained based on the ImageNet dataset~\cite{ImageNet}. For each semantic part, we used \cite{FastRCNN} to fine-tuned the VGG-16 using part annotations and obtained a specific part detector. In order to detect small object parts, we decreased the threshold for region proposal module and thus received more than 200 region proposals from each object region. For the YOLOv3 baseline, we used part annotations to fine-tune the pre-trained network.

The ninth baseline was a method for interactive annotating and learning object parts, which was proposed in \cite{ActivePart}. We called it \textit{Interactive-DPM}. The idea of online interactive learning of object parts is quite close to our method.

\textbf{Evaluation metrics:} We used two ways to evaluate part localization performance. The first metric is the APP~\cite{APPFerrari} (Average Percentage of Parts that are correctly estimated). Given each true object, we used the best pose/viewpoint component in the model (with the highest score) to explain the object. Then, for each object part of the pose/viewpoint, we used the ``$IOU\!>\!50\%$'' criterion~\cite{WeaklyDPM,SSDPM} to identify correct part localizations. We computed such a percentage for each type of semantic parts, and APP is the average for all the part types. To reduce the effects of object detection on the APP, we detect the object within the image region of $[c_{w}\!\pm\!w]$ and $[c_{h}\!\pm\!h]$, where $w$/$h$/$(c_{w},c_{h})$ indicates the width/height/center of the true object bounding box.

\begin{figure*}[th]
\centering
\includegraphics[width=1.0\linewidth]{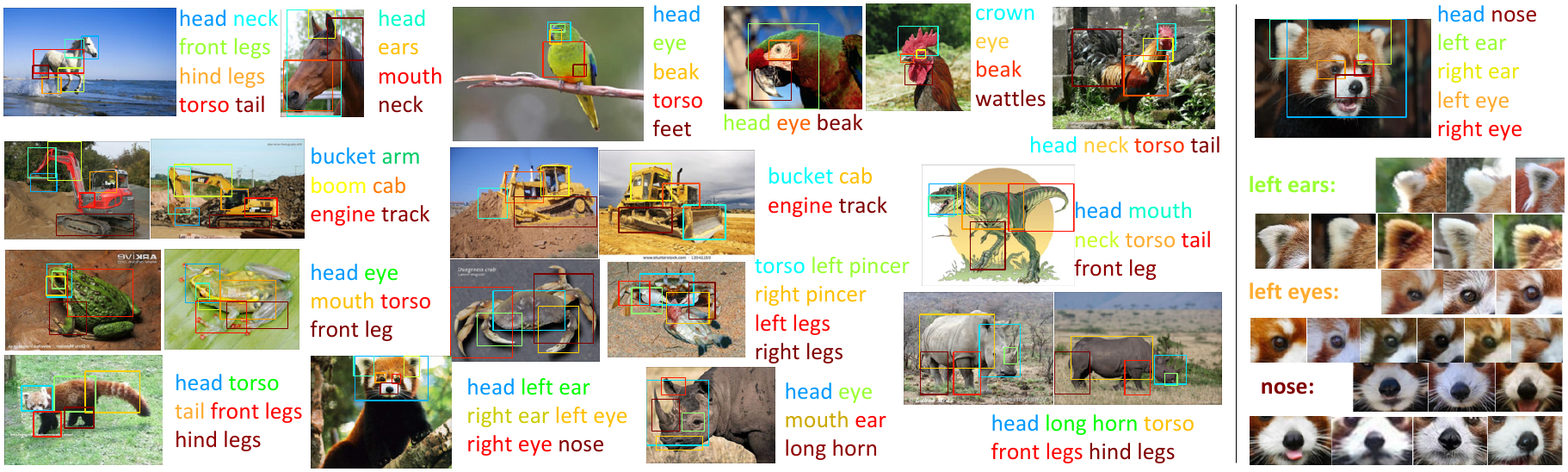}
\caption{AOG-based part localization. (Left) Explanation of parts in Layer 4. (Right) Semantic parts that are detected from different objects.}
\label{fig:results}
\end{figure*}

The second evaluation metric is the AER (average explanation rate) of objects. When an object is detected\footnote[9]{To simplify the evaluation metric, we only detected the best object from an image and ignored the others.}, if more than $2/3$ of the parts in its pose/viewpoint component are correctly localized, we consider this object being correctly explained by this component. Fig.~\ref{fig:compare_all} compares part localization performance between different baselines given a certain annotation cost. Different curves/dots correspond to different baselines. For most baselines, the annotation cost is the number of labeled parts on training samples. However, for our QA system, the overall cost consists of the cost of labeling parts and that of making yes/no judgments. Therefore, we drew two curves for our method: \textit{Ours} simply used the number of part boxes as the cost, whereas \textit{Ours (full cost)} computed the cost as {$(\# of boxes)\!+\!0.2\times(\# of judgements)$} (a judgment costs about $1/5$ of the time of labeling a part).

\begin{figure*}
\centering
\includegraphics[width=0.9\linewidth]{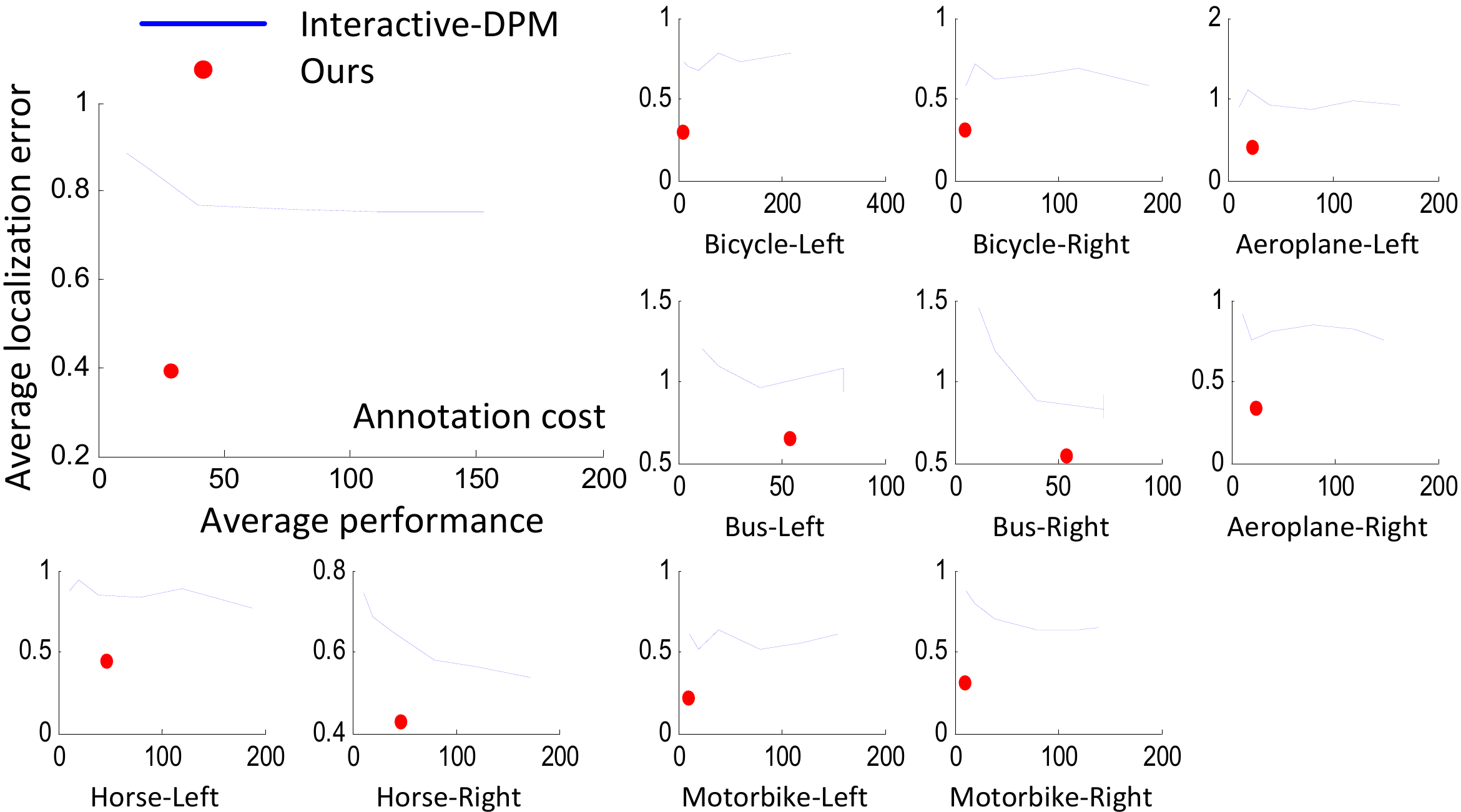}
\caption{Comparisons with \textit{Interactive-DPM} in terms of average localization errors. The annotation cost (horizontal axis) is computed based on \textbf{part} annotations. Our method exhibits low localization errors, given a limited number of part annotations.}
\label{fig:avgDist}
\end{figure*}

Note that the baseline of \textit{Interactive-DPM}~\cite{ActivePart} cannot detect bounding boxes for object parts, but localizes the center of each part. Therefore, just as in \cite{AverageLocalizationError}, we used the ``average localization error'' to evaluate the part localization accuracy. We normalized pixel error with respect to the part size, computed as $\sqrt{part\;height^2+part\;width^2}$. In Fig.~\ref{fig:avgDist}, we compared the proposed method with \textit{Interactive-DPM}~\cite{ActivePart} in terms of the average localization error.

\textbf{Comparison of learning efficiency.} We used the ten category models in the \textit{AOG-10} to explain its corresponding objects. For each category, 75 images were
prepared as testing images to compute the object explanation rate. Fig.~\ref{fig:results} illustrates part localization performance of the \textit{AOG-10}. Fig.~\ref{fig:compare_all} shows the average explanation rate over the ten categories. To evaluate our method, we computed the performance of intermediate models for each category, which were trained during the QA procedure with different numbers of storylines/questions. Given the same amount of labeling, our method exhibited about twice explanation rate of \textit{Matching}. When our method only used 125 bounding boxes for training, \emph{i.e.} $3\%$ of \textit{SSDPM-3}'s annotation cost (4258 boxes), it still achieved higher explanation rate than \textit{SSDPM-3} ($22.4\%$ vs. $17.9\%$).

\textbf{Performance on the Pascal VOC2007:} We learned the \textit{AOG-7} from web images, and tested it using \textit{horse}, \textit{motorbike}, \textit{bus}, \textit{aeroplane}, and \textit{bicycle} images with the \textit{left} and \textit{right} poses/viewpoints. This subset of Pascal images have been widely used for weakly-supervised exploiting part structures of objects~\cite{WeaklyDPM,ChoDiscovery}. We compare our method with the baselines of \textit{SSDPM}, \textit{P-Graph}, \textit{Fast-RCNN}, and \textit{Interactive-DPM}. \textit{SSDPM} used the Pascal training samples with the \textit{left} and \textit{right} poses/viewpoints for learning. We required \textit{SSDPM} to produce the maximum number of components for each category. Table~\ref{tab:VOC} shows the result. \textit{SSDPM} models were learned from different numbers of part annotations. In Fig.~\ref{fig:avgDist}, we compared the average localization errors of \textit{Interactive-DPM}~\cite{ActivePart} and our method.

\textit{SSDPM} used part annotations for training, so its performance depended on whether or not this method could extract discriminative features from small part regions. Therefore, \textit{SSDPM} may exhibit bad performance when the annotated parts were not distinguishable enough. In contrast, besides semantic parts, our method also mined discriminative latent parts from images, which increased the robustness of part localization. Unlike \textit{SSDPM}, \textit{P-Graph} and \textit{Interactive-DPM} directly learning part knowledge from a few annotations, we localized semantic parts on a latent object structure that was mined from unannotated web images. Thus, our method suffered less from the over-fitting problem. In addition, although \textit{Fast-RCNN} has exhibited superior performance in most object detection tasks, it did not perform that well in part detections. It is because 1) object parts were usually small in images, and without contextual knowledge, the low-resolution part patches could not provide enough distinguishing information; and 2) that we only annotated a small number of samples for each part (\emph{e.g.} $49/4=12.25$ annotations for each part of the aeroplane), which was not enough to learn a solid Fast-RCNN model. In contrast, our method did not require a large number of annotations for learning/fine-tuning, and modeled the spatial relationships between parts. Therefore, in Table~\ref{tab:VOC} and Fig.~\ref{fig:avgDist}, our method used fewer part annotations but achieved better localization accuracy.

\section{Conclusions and discussion}

In this study, we used human-computer dialogues to mine a nine-layer hierarchy of visual concepts from web images and build an AOG. Unlike the conventional problem of object detection that only focuses on object bounding boxes, our AOG localized semantic parts of objects and simultaneously aligned common shape primitives within each part, in order to provide a deep understanding of object statuses. In addition, our method combined QA-based active learning and weakly supervised web-scale learning, which exhibited high efficiency at knowledge mining in experiments.

In recent years, the development of the CNN has made great progress in object detection. Thus, it becomes more and more important to go beyond the object level and obtain a transparent understanding of deep object structures. Unlike widely used models (\emph{e.g.} CNNs for multi-category or fine-grained classification), the objective of our AOG model is not multi-category/fine-grained classification, but the deep explanation of the structural hierarchy of each specific object. We do not learn the AOG towards the application of multi-category classification. Instead, we design the loss for part localization and show the performance of hierarchical understanding of objects. Unlike object parts, the accuracy of detailed sketches within each local part is difficult to evaluate. Many of the sketches represent latent semantics within object parts.

Compared to deep neural networks, AOGs are more suitable for weakly-supervised learning of deep structures of objects. Figs.~\ref{fig:result_details_1} and \ref{fig:result_details_2} show one of the main achievements of this study, \emph{i.e.} the deformable deep compositional hierarchy of an object, which ranges from the ``object,'' ``semantic parts,'' ``sub-parts,'' to ``shape primitives.'' Such deep compositional hierarchy is difficult for deep neural networks to learn without given sufficient part annotations.

Our deep hierarchical representation of object structures partially solves the typical problem of how to define semantic parts for an object. In fact, different people may define semantic parts at different fine-grained levels. The uncertainty of part definition proves the necessity of our nine-layer AOG. Our AOG, for the first time, provides a nine-layer coarse-to-fine representation of object parts, which is a more flexible representation of object parts than shallow part models. People can define large-scale parts in the first four layers, and obtain representations of small parts in deep layers (please see Figs.~\ref{fig:result_details_1} and \ref{fig:result_details_2}). Thus, the flexibility of our AOG representation is one of main contributions of this research.

Although the AOG can be used for both object detection and part parsing, in recent years, deep neural networks~\cite{CNN,ResNet,denseNet} have exhibited superior the discrimination power to graphical models. Therefore, we believe the main value of the proposed method is weakly-supervised mining deep structure of objects, which can be used as explainable structural priors of objects for many applications and tasks. For example, a crucial bottleneck for generative networks is its limited interpretability. The automatically mined hierarchical object structures can be used as prior structural codes for generative networks and boost their interpretability.

The current AOG mainly models common part structures of objects without a strong discriminative power for fine-grained classification. However, our AOG can provide dense part correspondences between objects, which include both alignments of semantic parts and alignments of latent parts. Such dense part correspondences are crucial for fine-grained classification. More specifically, as discussed in \cite{parkAOG}, we can simply add different attributes to each node in the AOG. In this way, original AOG nodes mainly localize object parts, while attribute classifiers in AOG nodes servers for fine-grained classification.

Search engines usually return incorrect images without target objects and simple objects that are placed in image centers and well captured without occlusions. Thus, lifelong learning studies, such as \cite{Gpt_WeaklyCNN} and ours, mainly first learn from simple samples, and then gradually switch to difficult ones. In fact, comprehensive mining of all object poses/viewpoints, including infrequent poses/viewpoints, remains a challenging long-tail problem.

In this study, we aimed to explore a general QA system for model mining and test its efficiency. Thus, we applied simple features and trained simple classifiers for simplicity. However, we can extend the QA system to incorporate more sophisticated techniques (\emph{e.g.} connecting the AOG to the CNN) to achieve better performance. In experiments, we simply used very few (one or two) keywords for each category to search web images, because our weakly-supervised method did not need numerous web images for training. However, theoretically, people can also apply standard linguistic knowledge bases, such as WordNet~\cite{WordNet}, to provide several synonyms for the same category as keywords to search web images.

\section*{Acknowledgements}

This work is supported by ONR MURI project N00014-16-1-2007 and DARPA XAI Award N66001-17-2-4029, and NSF IIS 1423305.

\section*{Appendix: Objective function of graph mining}
\label{sec:append}

The objective function in \cite{OurICCV15AOG} was proposed in the form of
\begin{small}
\begin{equation}
\underset{{\boldsymbol\theta}_{PO_{i}}}{\arg\!\min}\Big\{\sum_{P\in Ch(PO_{i})}{\mathcal E}_{P}^{+}-\sum_{P\in Ch(PO_{i})}{\mathcal E}_{P}^{-}+\lambda\textrm{Complexity}({\boldsymbol\theta}_{PO_{i}})\Big\}\nonumber
\end{equation}
\end{small}
where the pattern complexity $\textrm{Complexity}({\boldsymbol\theta}_{PO_{i}})$ is formulated using the node number in the pattern, $\textrm{Complexity}({\boldsymbol\theta}_{PO_{i}})=\vert{Ch(PO_{i})}\vert+\beta\sum_{P\in Ch(PO_{i})}\vert{Ch(P)}\vert$. Then, the terms of ${\mathcal E}_{P}^{+}$ and ${\mathcal E}_{P}^{-}$ are the average responses of part node $P$ among positive images and negative images, respectively:
\begin{small}
\begin{equation}
\begin{split}
{\mathcal E}_{P}^{+}=E_{I\in\hat{\bf I}_{PO_{i}}}\Big\{S_{I}(P)+\underset{P'\in Ch(PO_{i}),P'\not=P}{\textrm{mean}}w_{PP'}S_{I}(P,P')\Big\}\\
{\mathcal E}_{P}^{-}=E_{I\not\in\hat{\bf I}_{PO_{i}}}\Big\{S_{I}(P)+\underset{P'\in Ch(PO_{i}),P'\not=P}{\textrm{mean}}w_{PP'}S_{I}(P,P')\Big\}\nonumber
\end{split}
\end{equation}
\end{small}
Considering $S_{I}^{app}(PO_{i})=0$, we can rewrite the objective as
\begin{small}
\begin{equation}
\begin{split}
&\underset{{\boldsymbol\theta}_{PO_{i}}}{\arg\!\min}\Big\{\sum_{P\in Ch(PO_{i})}\!\!\!{\mathcal E}_{P}^{+}-\sum_{P\in Ch(PO_{i})}\!\!\!{\mathcal E}_{P}^{-}+\lambda\textrm{Complexity}({\boldsymbol\theta}_{PO_{i}})\Big\}\\
=&\underset{{\boldsymbol\theta}_{PO_{i}}}{\arg\!\max}\Big\{\vert{Ch(PO_{i})}\vert\big\{\underset{I\not\in\hat{\bf I}_{PO_{i}}}{E}[S_{I}(PO_{i})]-\underset{I\in\hat{\bf I}_{PO_{i}}}{E}[S_{I}(PO_{i})]\big\}\\
&-\lambda\textrm{Complexity}({\boldsymbol\theta}_{PO_{i}})\Big\}\\
=&\underset{{\boldsymbol\theta}_{PO_{i}}}{\arg\!\max}\Big\{E_{I\in\hat{\bf I}_{PO_{i}}}[S_{I}(PO_{i})]-E_{I\not\in\hat{\bf I}_{PO_{i}}}[S_{I}(PO_{i})]\\
&-\frac{\lambda\textrm{Complexity}({\boldsymbol\theta}_{PO_{i}})}{\vert{Ch(PO_{i})}\vert}\Big\}\nonumber
\end{split}
\end{equation}
\end{small}
In addition, the average score of $S_{I}(PO_{i})$ for negative (background) images is normalized to zero. Therefore, we can further approximate the objective as
\begin{small}
\begin{equation}
\underset{{\boldsymbol\theta}_{PO_{i}}}{\arg\!\max}\Big\{E_{I\in\hat{\bf I}_{PO_{i}}}[S_{I}(PO_{i})]-\frac{\lambda\textrm{Complexity}({\boldsymbol\theta}_{PO_{i}})}{\vert{Ch(PO_{i})}\vert}\Big\}\nonumber
\end{equation}
\end{small}
Therefore, if we redefine a new complexity $\textrm{Complexity}^{new}({\boldsymbol\theta}_{PO_{i}})={\textrm{Complexity}({\boldsymbol\theta}_{PO_{i}})}/{\vert{Ch(PO_{i})}\vert}$, we can write the objective function as
\begin{small}
\begin{equation}
\underset{{\boldsymbol\theta}_{PO_{i}}}{\arg\!\max}\Big\{E_{I\in\hat{\bf I}_{PO_{i}}}[S_{I}(PO_{i})]-\lambda\textrm{Complexity}^{new}({\boldsymbol\theta}_{PO_{i}})\Big\}\nonumber
\end{equation}
\end{small}

{\small
\bibliographystyle{ieee}
\bibliography{TheBib}
}

\end{document}